\documentclass[letterpaper]{article} 
\usepackage{aaai25}  
\usepackage{times}  
\usepackage{helvet}  
\usepackage{courier}  
\usepackage[hyphens]{url}  
\usepackage{graphicx} 
\usepackage{amsfonts}
\usepackage{hyperref}
\hypersetup{hidelinks,
	colorlinks=true,
	allcolors=black,
	pdfstartview=Fit,
    urlcolor=blue,
	breaklinks=true}
\usepackage{url}

\usepackage{natbib}  
\usepackage{caption} 
\usepackage{algorithm}
\usepackage{algorithmic}

\usepackage{xcolor}

\usepackage{newfloat}
\usepackage{listings}
\DeclareCaptionStyle{ruled}{labelfont=normalfont,labelsep=colon,strut=off} 
\floatstyle{ruled}
\newfloat{listing}{tb}{lst}{}
\floatname{listing}{Listing}
\title{MyGo: Consistent and Controllable Multi-View Driving Video Generation \\with Camera Control}
\author{
    Yining Yao\textsuperscript{\rm 1}, Xi Guo\textsuperscript{\rm 2}, Chenjing Ding\textsuperscript{\rm 2}, Wei Wu\textsuperscript{\rm1,\rm 2}\thanks{Corresponding author.}
}

\affiliations{
    \textsuperscript{\rm 1}Tsinghua University\\
    \textsuperscript{\rm 2}Sensetime Research\\
    yaoyn23@mails.tsinghua.edu.cn\\
    \{guoxi,dingchenjing,wuwei\}@sensetime.com

}

\usepackage{bibentry}

\begin{document}

\maketitle
\begin{figure*}[t]
\centering
\includegraphics[width=1.96\columnwidth]{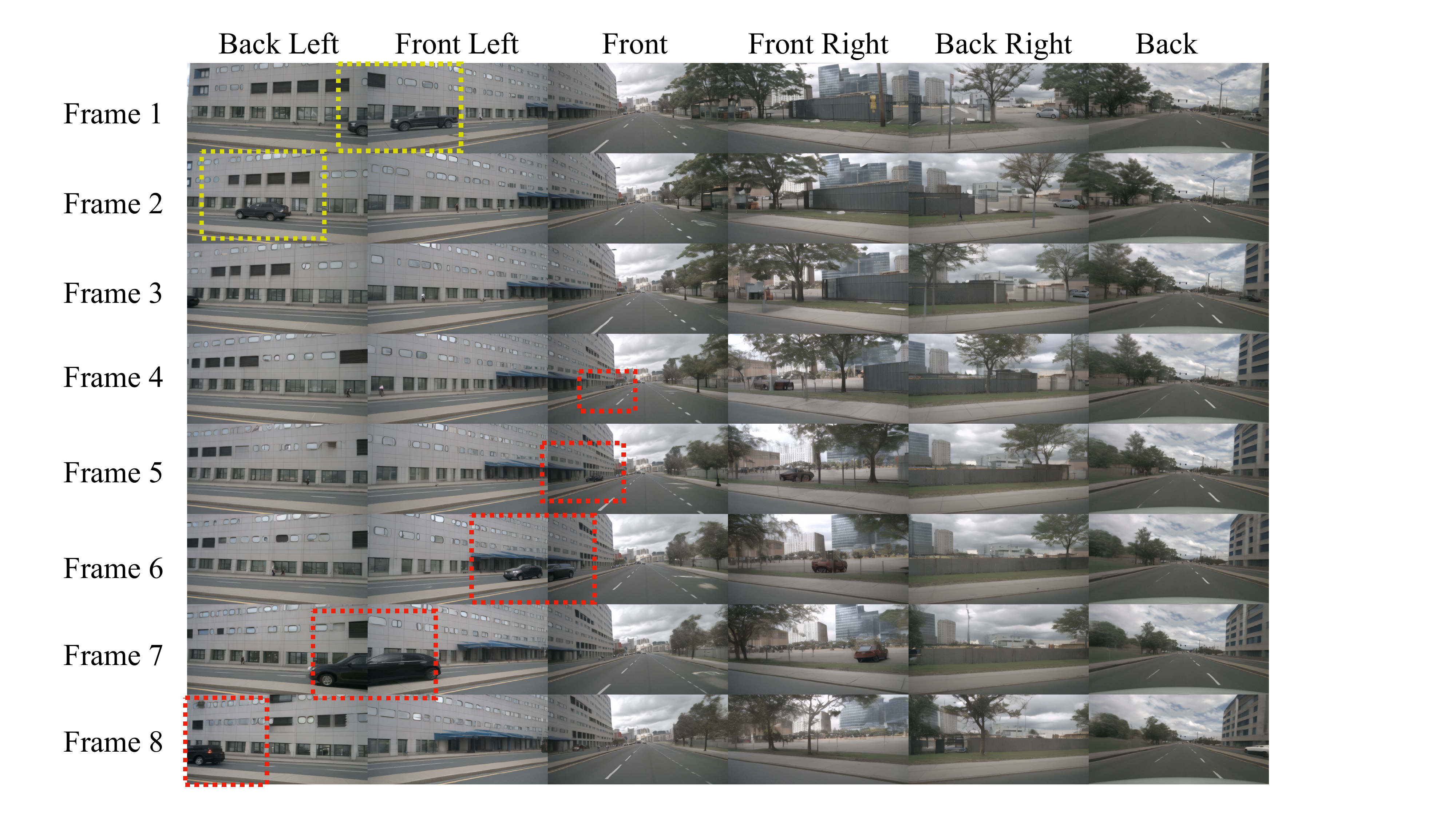} 
\caption{Examples of the generated multi-view video frames. MyGo is capable of generating multi-view videos precisely controlled by onboard camera parameters as well as road structural information while maintaining excellent temporal consistency as well as cross-view long-term spatial consistency.
}
\label{main_result}
\end{figure*}

\begin{abstract}
High-quality driving video generation is crucial for providing training data for autonomous driving models. However, current generative models rarely focus on enhancing camera motion control under multi-view tasks, which is essential for driving video generation. Therefore, we propose MyGo, an end-to-end framework for video generation, introducing motion of onboard cameras as conditions to make progress in camera controllability and multi-view consistency. MyGo employs additional plug-in modules to inject camera parameters into the pre-trained video diffusion model, which retains the extensive knowledge of the pre-trained model as much as possible. Furthermore, we use epipolar constraints and neighbor view information during the generation process of each view to enhance spatial-temporal consistency. Experimental results show that MyGo has achieved state-of-the-art results in both general camera-controlled video generation and multi-view driving video generation tasks, which lays the foundation for more accurate environment simulation in autonomous driving. Project page: \url{https://metadrivescape.github.io/papers_project/MyGo/page.html}
\end{abstract}

%
\section{Introduction}
With the advancements in deep learning, autonomous driving has garnered more research both in academia and industry. In the perception and decision processes of the autonomous driving system, many studies utilize multi-view video from onboard cameras to generate geometrically structured bird’s-eye view (BEV) representations to depict traffic conditions \cite{li2022bevformer, yang2023bevformer}. However, acquiring a substantial amount of multi-view driving video data is challenging due to the high costs of collection and annotation. Therefore, research on generating high-quality driving videos with multi-view consistency and vehicle motion controllability is of significant value for the development of autonomous driving systems.

In the field of video generation, diffusion models have demonstrated promising capabilities in generating new videos that align closely with real-world data distributions \cite{ho2020denoising,dhariwal2021diffusion}. Moreover, studies have confirmed that samples generated by diffusion model can assist downstream tasks such as representation extraction and segmentation \cite{swerdlow2024street,gao2023magicdrive,yang2023bevcontrol}. Based on the works of unconditional generative models, studies on controllable generation using text, images, and other modalities have also shown impressive results, indicating that applying diffusion models to multi-view controllable driving video generation presents a valuable research opportunity \cite{zhang2023adding,blattmann2023stable,ramesh2021zero}.

For multiview driving video generation, controllability and consistency are two crucial objectives. Controllability refers to the ability to edit the street scene style, as well as the motion of the ego vehicle and other vehicles or pedestrians in the video, using corresponding conditioning inputs.For controlling street scene style, information can typically be provided through keyframes in an image-to-video generative model. The 3D object bounding boxes and road map contain motion information of other vehicles . Onboard camera pose parameters offer sufficient information to control the motion of ego vehicle. Additionally, annotated text can provide finer control over conditions such as weather, lighting, and other factors. Consistency means that the generated video should maintain coherence across both the temporal dimension and across the spatial dimension at different camera positions at the same frame. Temporal consistency can often be effectively controlled using the temporal transformer block in diffusion U-Net. However, spatial consistency in multi-view video generation tasks, particularly in controlling each view according to its corresponding camera pose, still lacks comprehensive study.

Previous studies of driving video generation have primarily focused on using layout to control the generation process, such as in \cite{gao2023magicdrive,yang2023bevcontrol,swerdlow2024street,zheng2023layoutdiffusion,cheng2023layoutdiffuse}. However, these studies mainly focus on controlling the motion of other objects in the scene, lacking controllability of the motion of the ego vehicle, which is also of significant importance for ensuring the controllability of generation. The rotation and translation information of onboard camera can be seen as the movement information of ego vehicle, which can be used to guide the generation process. On the other hand, current models enabling camera pose control primarily focus on the following aspects: 1) represent camera parameters in a way that can be better learned by neural networks, and 2) incorporate camera pose conditions into pre-trained generative models, yet research on camera control within the context of consistent multi-view generation tasks remains underexplored\cite{xu2024camcocameracontrollable3dconsistentimagetovideo,he2024cameractrlenablingcameracontrol,motionctrl,hou2024trainingfreecameracontrolvideo}. Some studies directly concatenate or add camera pose to the input of the temporal attention layer in the pre-trained U-Net\cite{he2024cameractrlenablingcameracontrol,motionctrl}. However, this method is not conducive to preserving the rich knowledge learned in the pre-train process and also ignore effective coordination with other control conditions.

To overcome the above issues, we purpose MyGo, an end-to-end framework that generates multi-view driving video with good controllability and consistency, using the pose of onboard camera to control the movement of ego car. Inspired by the structure of ControlNet, we introduce a camera control module to SVD U-Net, which retains the extensive knowledge of the pre-trained model and achieves effective motion-editable video generation. To apply camera control to multi-view generation tasks, we use camera pose based epipolar geometry to constrain neighbour camera view condition and further guide multi-view video generation, which helps to achieve spatial consistency across different viewpoints.

To summarize, the main contributions of this work are:
\begin{itemize}
    \item We introduce MyGo, an end-to-end framework that generates multi-view driving video with good controllability and consistency.
    \item The ControlNet-like camera control module retains the extensive knowledge of the pre-trained model and achieves effective motion-editable video generation.
    \item We use camera pose based epipolar geometry to constrain neighbour camera view condition producing multi-view video with spatial consistency.
    \item We achieve state-of-the-art video synthesis performance on nuScenes dataset, and also attained the highest camera control accuracy on RealEstate10K dataset.
\end{itemize}

\section{Related Works}
\subsection{Multi-view Video Generation}
In the research on multi-view video generation, temporal and spatial consistency are crucial objectives. To keep multi-view consistency, CVD\cite{cvd} applies a masked cross-view attention between the frames from the two videos. MVDiffusion\cite{deng2023mv} proposes a correspondence-aware attention module to align the information from multiple views. Vivid-Zoo\cite{li2024vividzoomultiviewvideogeneration} designs 2D-3D and 3D-2D alignment module to connect pre-trained 2D video and multi-view image diffusion model. However, there is an absence of effective mechanisms for utilizing viewpoint position parameters in those methods. Therefore, we not only purpose a camera control module integrated into the U-Net downsample and upsample blocks, but also introduce a crossview condition module that uses epipolar geometry based attention mask to combine camera pose conditions and crossview conditions.
\subsection{Camera Controlled Video Generation}
Many studies of camera controlled video generation focus on how to incorporate camera condition into pre-trained models. MotionCtrl\cite{motionctrl} simply repeats the rotation and translation of camera, which is further appended to the latent in temporal transformer block. CameraCtrl\cite{he2024cameractrlenablingcameracontrol} takes the plücker embedding as input, and is add to the latent features. Some other studies use epipolar geometry between cameras as a condition. CamCo\cite{xu2024camcocameracontrollable3dconsistentimagetovideo} purposes epipolar constraint attention block, gathering information from the corresponding region of the first frame. Different from studies described above, the camera control module in this paper is not directly integrated into the input of the pre-trained model. Instead, it utilizes additional connections to achieve more precise control.

\section{Method}
\subsection{Preliminary}
Diffusion models \cite{dhariwal2021diffusion,ho2020denoising,rombach2022high,ramesh2021zero,zhang2023adding,song2020denoising} are a family of probabilistic generative models that are trained to learn a data distribution by progressively removing a variable (noise) sampled from an initial Gaussian distribution. With the noise gradually adding to $z$ for $t$ steps, a latent variable $z$ and its noisy version $z_t$ will be obtained, the objective function of latent diffusion models can be simplified to
\begin{equation}
\mathbb{E}_{z,c,\epsilon\sim{N}(0,1),t}\left [ \left \Vert 
\epsilon - \epsilon_\theta(z_t,t)
\right \Vert_2^2 \right ] 
\end{equation}
which is the squared error between the added noise $\epsilon$ and the predicted noise $\epsilon_\theta(z_t,t)$ by a neural model $\epsilon_\theta$ at time step $t$, given $c$ as condition. This approach can be generalized for learning a conditional distribution, the network $\epsilon_\theta(z_t, t, c)$ can faithfully sample from a distribution conditioned on $c$. In this work, we leverage a pre-trained image-to-video Latent Diffusion Model (LDM)\cite{rombach2022high}, i.e., Stable Video Diffusion, in which the whole diffusion process has proceeded in the latent space of a pre-trained VAE\cite{vae} with an image encoder and a spatial-temporal decoder.

\begin{figure*}[t]
\centering
\includegraphics[width=1.8\columnwidth]{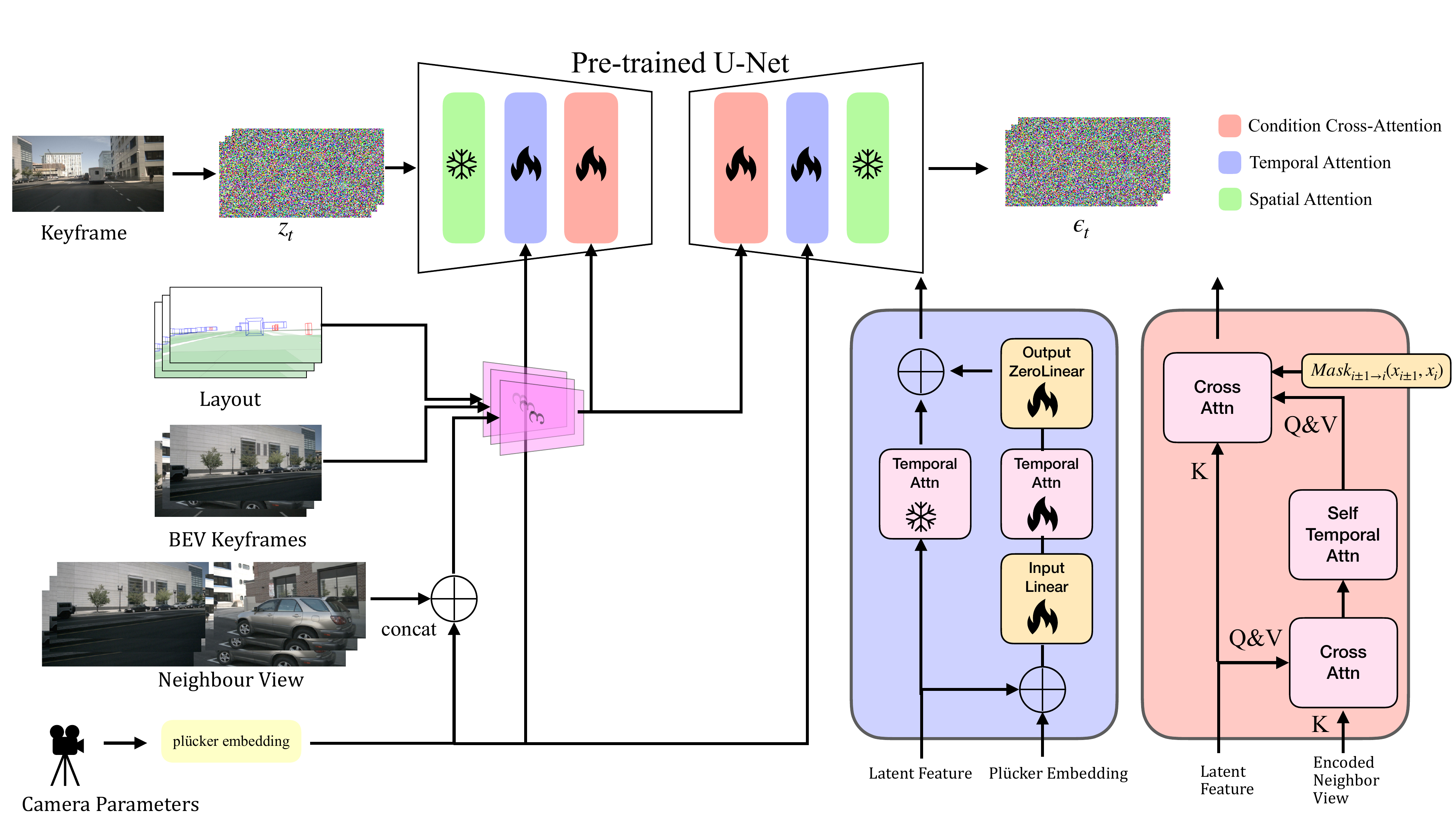} 
\caption{MyGo takes BEV map, 3D bounding boxes, neighbour view, keyframes and camera parameters as conditions, and uses a unified encoder to process the conditions. The encoded conditions are further integrated into U-Net by a condition cross-attention block. We design a ControlNet like structure to 
inject camera plücker coordinates into pre-trained U-Net blocks. Moreover, in neighbour view cross-attention block, we use epipolar geometry as a constraint to guide the calculation of cross-attention.}
\label{overview}
\end{figure*}

\subsection{Overview}
The overview of MyGo is illustrated in Fig \ref{overview}. Based on the pipeline in the official Diffusers codebase\cite{von2022diffusers}, MyGo utilizes the intrinsic and extrinsic parameters of cameras mounted on the ego car, along with pixel-level conditions such as the BEV map, 3D bounding boxes, the first frame of each view, and neighbor view videos as input conditions to generate high-quality multiview driving videos. For pixel-level conditions, we employ a unified bidirectional modulated spatial-temporal transformer structure to process them. To integrate camera conditions and control the ego car's movement, we first use plücker embedding \cite{plucker1828analytisch} to map the camera parameters to the shape of the pixel-level conditions, which is detailed in Section \ref{subsec:3.3}. A camera control module that integrates camera pose into the pre-trained blocks of the video generator will be introduced in Section \ref{subsec:3.4}. Furthermore, we propose a camera pose guided neighbor view transformer block to enhance multipview consistency in Section \ref{subsec:3.5}.
\subsection{Representing Camera Condition}\label{subsec:3.3}
To enable our model to effectively learn the real-world meaning of camera parameters, it is crucial to select a better representation for modeling the camera. In most datasets, camera parameters are represented in their raw form as extrinsics $\mathbf{E}$ and intrinsics $\mathbf{K}\in\mathbb{R}^{3\times3}$, where $\mathbf{E} = [\mathbf{R};\mathbf{t}]$ includes the camera's rotation $\mathbf{R}\in\mathbb{R}^{3\times3}$ and translation $\mathbf{t}\in\mathbb{R}^{3\times1}$ relative to the world coordinate system, and $\mathbf{K}$ represents the mapping from the camera coordinate space to the image coordinate space. However, to facilitate the combination of camera parameters with pixel-level latent and conditions, we need to further map the camera parameters to the pixel-level space. Some studies \cite{gao2023magicdrive,motionctrl} simply repeat or use neural network to process camera conditions, without considering that the camera parameters themselves physically have already implied pixel level information. Additionally, the mismatches in the numerical scales of $\mathbf{R}$, $\mathbf{t}$, and $\mathbf{K}$ can also introduce difficulties in the learning process.

Considering the drawbacks of the aforementioned methods, MyGo uses plücker embedding to preprocess the camera parameters, facilitating subsequent integration into the model. For pixel points $(u,v)$ at the target resolution $h \times w$, the plücker coordinates can be represented as $\mathbf{p}_{u,v} = (\mathbf{t} \times \mathbf{d}_{u,v}$), where $\mathbf{d}_{u,v} = \mathbf{RK^{-1}}(u,v,1)^T + \mathbf{t}$, representing the direction of the ray from the camera center to the image plane. The plücker coordinate of camera parameters is further adjusted to the coordinate space relative to the initial position of the camera.

\subsection{Integrate Camera Pose into Video Generator}\label{subsec:3.4}
After obtaining the plücker coordinates $\mathbf{P}\in\mathbb{R}^{{n}\times6\times{h}\times{w}}$ of a n-frame video, to achieve precise and high-quality camera control, it is necessary to inject the plücker coordinates into the U-Net network of the diffusion model. In our method, we divide this process into two parts: downsampling the plücker coordinates to different resolutions of the U-Net and integrating the processed plücker coordinates into the U-Net.

Firstly, we sequentially stack convolution layers and ResNet blocks to map the pixel-level plücker coordinates to different resolutions of the U-Net output, which can be expressed as $ \mathbf{P} = [\mathbf{p}_1, ..., \mathbf{p}_i], \mathbf{p}_i\in\mathbb{R}^{{n}\times6\times{h}_i\times{w}_i}$. This encoder architecture is also applied to other conditions such as BEV maps and bounding boxes. Using the same architecture is important to transform different modalities into a unified representation, facilitating their subsequent incorporation.

To control the ego vehicle's motion with camera condition, we incorporate $ \mathbf{P}$ into the pre-trained downsample and upsample blocks. Inspired by ControlNet\cite{zhang2023adding}, we design a method that decouples camera control from high-quality video generation and avoids unnecessary interference between the two. As shown in Fig \ref{overview}, in each spatial-temporal transformer block, we integrate camera conditions before the temporal transformer, generating movement that aligned with camera poses among the frames. Denote the latent feature which is input of the temporal transformer as $\mathbf{z}_i$, and the corresponding camera condition as $\mathbf{p}_i$. We first concatenate $\mathbf{p}_i$ to $\mathbf{z}_i$, and use a linear layer whose weights corresponding to $\mathbf{z}_i$ are set as an identical mapping while weights corresponding to $\mathbf{p}_i$ are zero-initialized. \begin{equation}\label{eq1}\mathbf{z}_i^{cam} = Linear_{in}(concat(\mathbf{z}_i,\mathbf{p}_i))\end{equation}Then we use a temporal transformer block same to the original one to process the output of the linear layer and further pass the output through another zero-initialized linear block. 
\begin{equation}\label{eq2}\mathbf{z}_i^{cam} = Linear_{out}(TemporalAttn(\mathbf{z}_i^{cam}))\end{equation}
Including camera condition, $\mathbf{z}_i^{cam}$ is added to the output of the pre-trained temporal transformer.
\begin{equation}
\mathbf{z}_i^{out} = \mathbf{z}_i^{cam} + TemporalAttn(\mathbf{z}_i)  
\end{equation}
While training, we freeze the parameters of original transformer to enhance the model to focus on the learning of camera control. It is important to initialize the weights of camera control module with zero, which maintains the video generation capabilities of the pre-trained model while enabling more independent and precise control of camera motion.

\begin{figure}[ht]
\centering
\includegraphics[width=0.95\columnwidth]{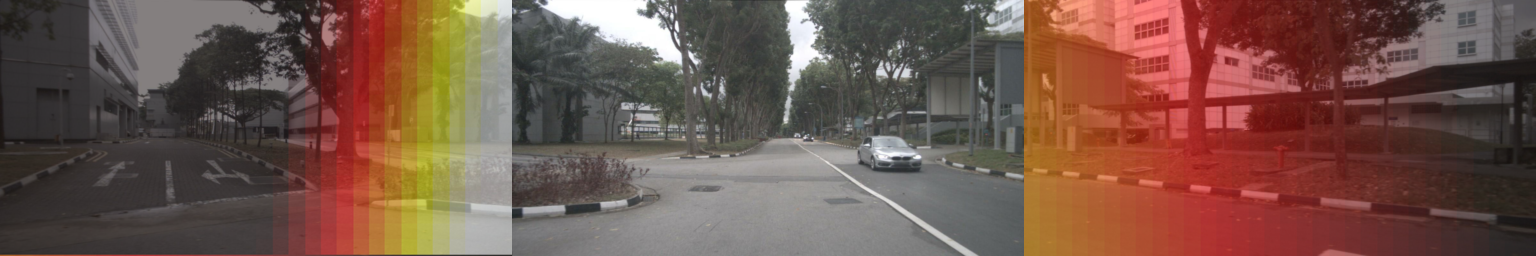} 
\includegraphics[width=0.35\columnwidth]{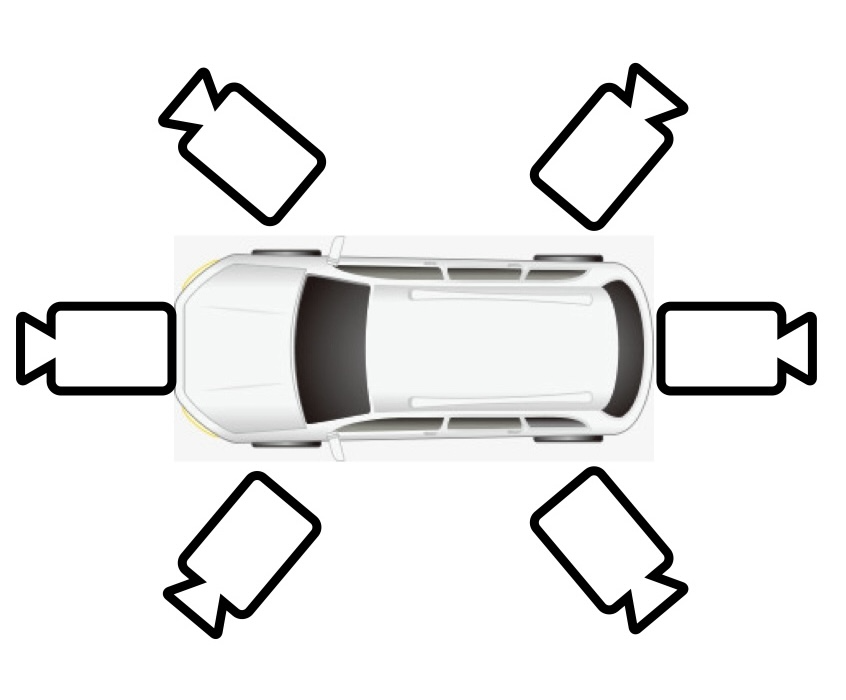} 
\caption{Visualization of attention mask based on epipolar geometry. During the  neighbour view cross-attention process, a mask is computed so that the right side of the left neighbor and the left side of the right neighbor are included in the calculation, while other parts are ignored}
\label{mask}
\end{figure}
\begin{figure*}[ht]
\centering
\includegraphics[width=1.95\columnwidth]{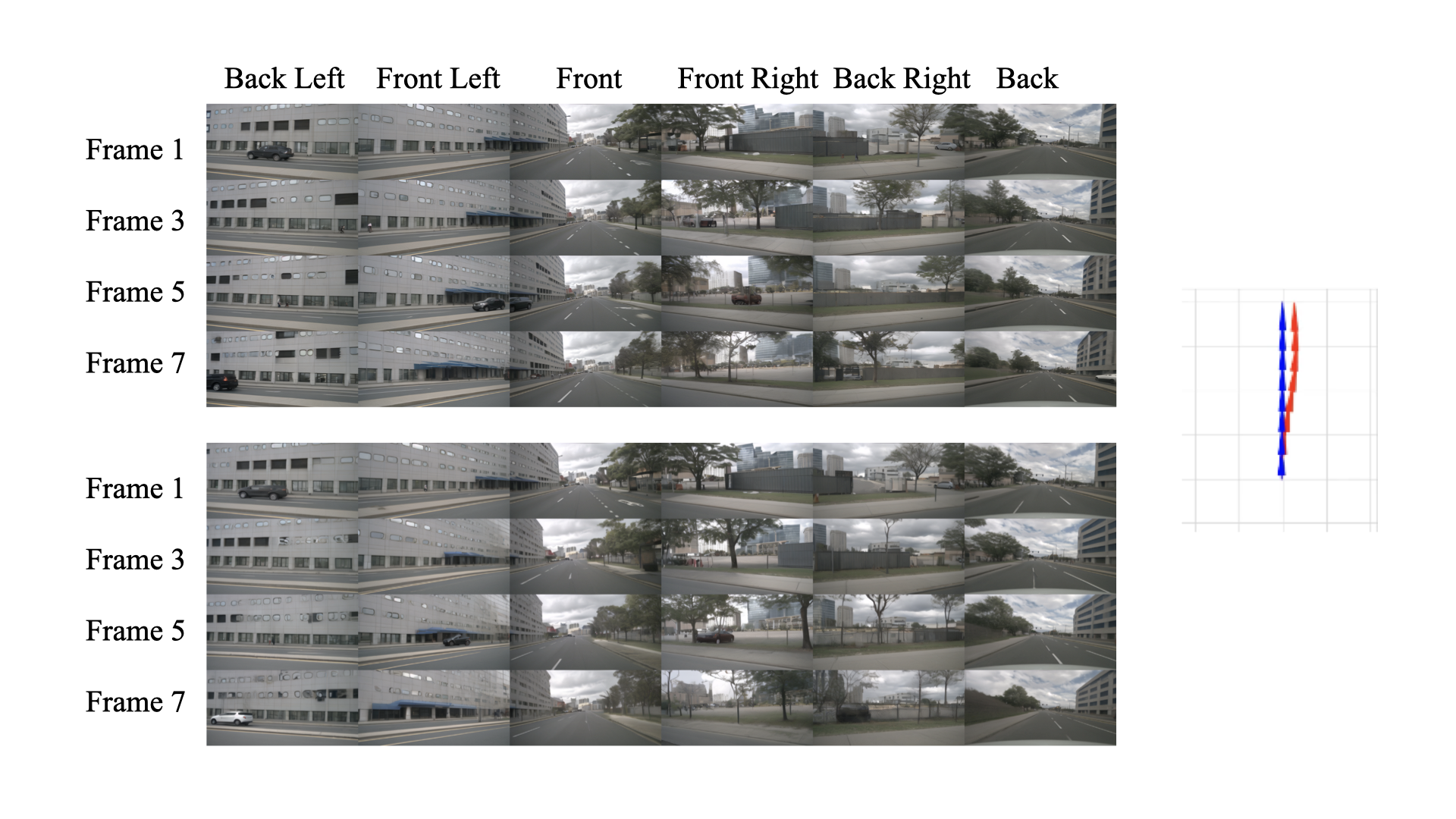} 

\caption{Result of the case of changing to another line, demonstrating that our method can edit the ego vehicle's motion while maintaining generation quality and spatial-temporal consistency}
\label{newroute}
\end{figure*}
\subsection{Use Neighbor View Information with Camera Condition to Enhance Multi-view Consistency}\label{subsec:3.5}
In multi-view video generation task, achieving multi-view consistency with camera control is a crucial objective. Based on the unified cross-attention structure that are used to combine pixel-level condition and latent feature, we purpose a cross view mechanism to combine videos of neighbour camera and the camera poses as shown in Fig \ref{overview}. 
Considering onboard cameras are usually placed around the ego vehicle, there is always a set of cameras with almost no discernible overlapping regions between them; in other words, the content captured by these cameras is largely independent of each other. Therefore we first generate video of the set of cameras, and further use them as neighbour view conditions, and integrate them with camera conditions to guide the synthesis of videos from other viewpoints. This approach allows the generative model to directly access content from other relevant viewpoints, thereby avoiding inconsistencies between views and improving the quality of the generated output. 

Specifically, the plücker coordinates of local and neighbour cameras are first concatenate to corresponding neighbour view videos, with a size of $\mathbb{R}^{{(c+6+6)}\times{h}\times{w}}$ where $c$ is the channel of neighbour view videos and usually is 3. The concatenated condition tensor are encoded by the module mentioned in \ref{subsec:3.4} to downsample to various resolution.

Generally, every pixel level condition including neighbour view videos, BEV map, 3D bounding box are processed by corresponding cross-attention block that is plug into the U-Net before every downsample and upsample block. However,  due to only partial overlap between adjacent viewpoints, the extraneous information might interfere with spatial consistency in neighbour view cross-attention block. Therefore we use epipolar geometry to guide the cross-attention process and force the network only focus on the relevant part.
Denote the raw camera parameters of local camera and its neighbour as $[cam_i,cam_{i-1},cam_{i+1}$], we can calculate the fundamental matrix between each two cameras $F_{i\pm1\to{i}}$. According to epipolar geometry constraint , denote the projection of a 3D point $\mathbf{P}$ in the real world onto the image planes of camera 1 and camera 2 as $x_1=(u_1, v_1),x_2=(u_2, v_2)$, 
\begin{equation}
(u_2, v_2)F_{1\to{2}}(u_1, v_1)^T = 0
\end{equation}
Therefore, we select the most important pixel points of neighbour views and mask out other irrelevant points following:
\begin{equation}
Mask_{i\pm1\to{i}}(x_{i\pm1},x_i) = \mathbf{1}(x_iF_{i\pm1\to{i}}x_{i\pm1}^T < \tau)
\end{equation} where $\tau$ is the first quartile of $x_iF_{i\pm1\to{i}}x_{i\pm1}^T$
And we apply $Mask_{i\pm1\to{i}}$ to the cross-attention module as illustrated in Fig \ref{mask}, which only select the points that overlap with the current viewpoint to enhance multi-view consistency.

\begin{table*}[t]
\centering
\begin{tabular}{c|c|c|c|c|c}
\hline              
 \textbf{Dataset}         & \textbf{Model}   & FID↓ & FVD↓ & RotErr↓& TransErr↓\\ \hline
 & MotionCtrl& 12.66 & 171.10& 9.32&10.34\\ 
                      {RealEstate10K}      &CameraCtrl& \textbf{11.18}   & \textbf{152.01} & 8.36&6.27\\ 
                             &\textbf{MyGo} (Ours) &  12.84  &   162.83    &   \textbf{7.71}&\textbf{5.98}\\  \hline
\end{tabular}
\caption{The quantitative results on RealEstate10K dataset. The comparative analysis of our method against others is detailed in Sec. \ref{subsec:4.2}. The top-performing results have been highlighted in \textbf{bold} for clarity and emphasis.}
\label{tab:metric1}
\end{table*}

\begin{table*}[t]
\centering
\begin{tabular}{c|c|c|c|c}
\hline             
 \textbf{Dataset}         & \textbf{Model}   & FID↓ & FVD↓ & KPM Score↑  \\ \hline
     &  MotionCtrl& 12.94 & 135.17&80.83\%\\ 
                     {nuScenes}         &CameraCtrl&12.65   & 126.15 &80.67\%  \\ 
                             &\textbf{MyGo} (Ours) &  \textbf{11.60} &\textbf{122.06}   &\textbf{81.22\%}           \\  \hline
\end{tabular}
\caption{The quantitative results on nuScenes dataset.}
\label{tab:metric2}
\end{table*}

\section{Experiments}
\subsection{Experiment Details}\label{subsec:4.1}
\subsubsection{Datasets}
The training data is sourced from the real-world driving dataset nuScenes \cite{caesar2020nuscenes}, a prevalent dataset in BEV segmentation and 3D object detection for driving scenarios. We follow the official settings, utilizing 700 street-view scenes for training and 150 for validation. Our method considers 8 object classes and 8 road classes. To further validate the capabilities of the camera control module, we also train our model on the RealEstate10K dataset \cite{real10k}, which is widely used in studies of single-view camera control generation. On the RealEstate10K dataset, we use all samples from training set for training and randomly sampled 1000 samples for evaluation.
\subsubsection{Evaluation Metrics}
To comprehensively assess the effectiveness of our approach in achieving realism, consistency, and controllablity, we have selected four key metrics for comparison with related video generation methods. To evaluate realism, we use the well-established Fréchet Inception Distance (FID) \cite{fid} to measure the quality of the synthesized images. For ensuring that our videos maintain consistent and smooth motion, we employ the Fréchet Video Distance (FVD) \cite{fvd} as an indicator of temporal coherence. To evaluate the degree to which the motion in the generated video aligns with the camera condition, we run COLMAP \cite{colmap} pose estimation on the generated videos of RealEstate10K dataset, and compare the errors between the estimated camera rotation and translation with the ground truth separately. Considering that for samples with poor camera controllability, COLMAP always fails to match video frames and complete the estimation, we incorporate the success rate of COLMAP estimates as a weight in the error calculation:
\begin{equation}
    RotErr = \frac{1}{SuccessRate}\times\sum^n_{j=1}{arccos\frac{tr(\mathbf{R}_{gen}^j\mathbf{R}_{gt}^{jT})-1}{2}}
\end{equation}
\begin{equation}
    TransErr = \frac{1}{SuccessRate}\times\sum^n_{j=1}{\left \Vert \mathbf{T}^j_{gen} - \mathbf{T}^j_{gt} \right \Vert_2}
\end{equation}
where $SuccessRate$ is the ratio of the number of samples successfully matched by COLMAP and all generated samples, $\mathbf{R}_{gen}^j$ and $\mathbf{R}_{gt}^{jT}$ ,$\mathbf{T}^j_{gen}$ and $\mathbf{T}^j_{gt}$ are the rotation and normalized translation matrix relative to the first frame.
Following \cite{kpm}, we use the Key Points Matching (KPM) score, to evaluate multi-view consistency. This metric use a pre-trained model\cite{kpm1} to match key points and calculate the ratio of the number of matched points in generated data and ground truth.

\subsubsection{Implement Details}
We implement our approach based on the official codebase of Diffusers\cite{von2022diffusers}, and a pre-trained SVD video generation model\cite{blattmann2023stable}. For training and validation, we compiled images into numerous sequences of 8-frame videos on nuScenes and 14-frame on RealEstate10K, each of which is resized into a size of 256 × 512. All models are trained with the 8-bit AdamW optimizer with a base learning rate of 1e-5 and a batch size of 8. To preserve the capabilities of the pre-trained model, we employ zero initialization for the last layer of newly added modules. Additionally, we observed that using a temporal increasing classifier-free guidance scale from SVD leads to gradual distortion during iterative inference. Therefore, in all experiments, we fixed the scale at 2.5.
\subsubsection{Baselines}
We compare our results with two recent video generative model with camera control: CameraCtrl\cite{he2024cameractrlenablingcameracontrol} and MotionCtrl\cite{motionctrl}. Despite therr are other camera control generation methods such as CamCo\cite{xu2024camcocameracontrollable3dconsistentimagetovideo}, the models and codes are not open source. It should be noted that on single-view RealEstate10K dataset, we only take camera parameters as input, while ignore other conditions such as keyframes or BEV map. All models share the same hyperparameters and other settings.

\begin{figure}[H]
\centering
\includegraphics[width=0.95\columnwidth]{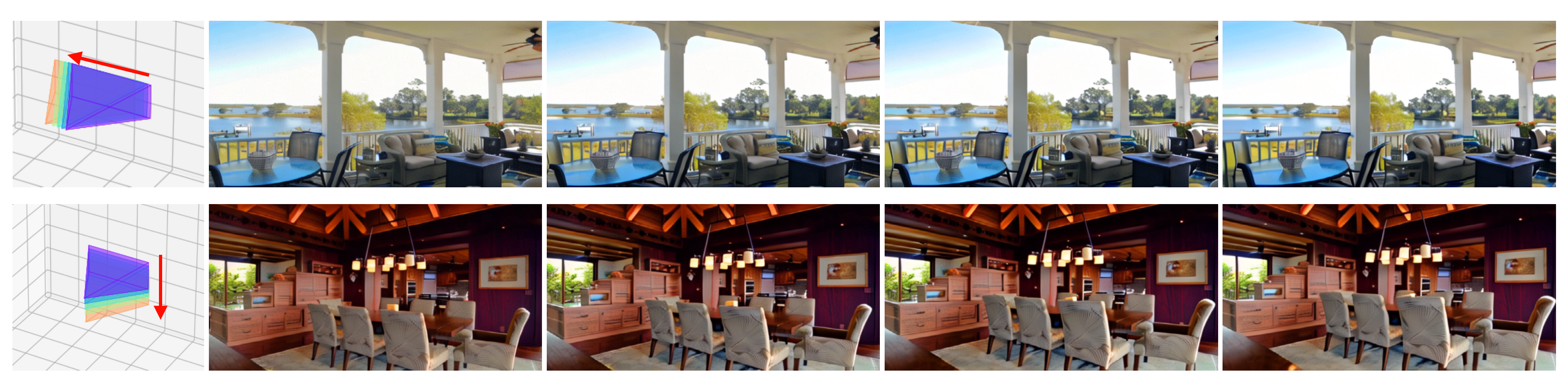} 
\includegraphics[width=0.95\columnwidth]{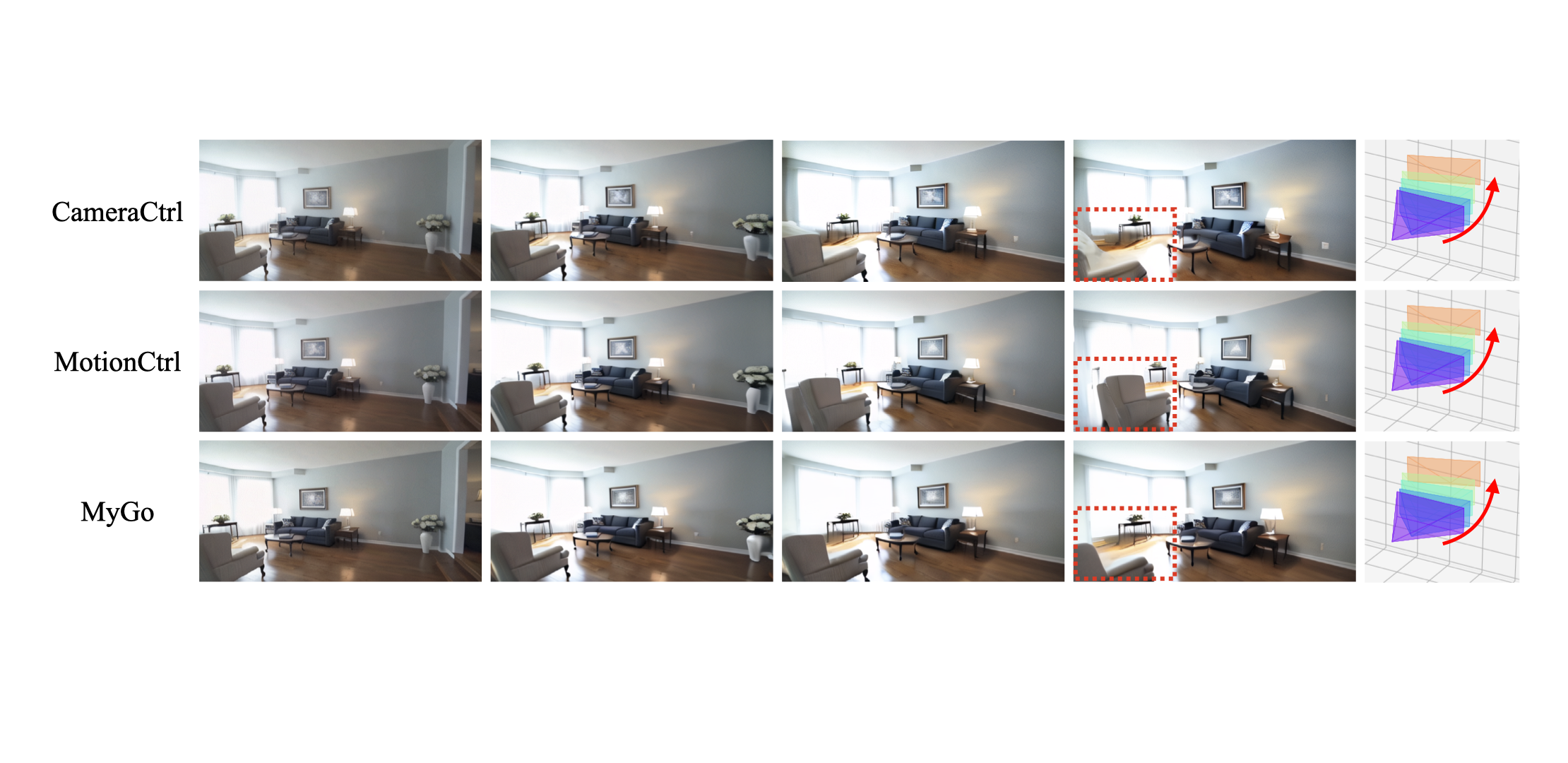} 
\caption{Experiments on RealEstate10K, including generation results of several camera  trajectories and comparison with baseline methods. Content in red boxes shows how our methods outperforms baselines in camera controllability.}
\label{real10k}
\end{figure}
\subsection{Experiment Results}\label{subsec:4.2}
\subsubsection{Quantitative Results}
Tab \ref{tab:metric1} and Tab \ref{tab:metric2} shows the quantitative results of experiments on RealEstate10K and nuScenes dataset. Due to the dynamic nature of scenes in the nuScenes dataset, COLMAP faces significant challenges in successfully matching them, we only verify camera controllability through $RotErr$ and $TransErr$ on realEstate10K dataset. 
Multiview driving video generation experiments conducted on the nuScenes dataset demonstrate that our method outperforms the baseline methods in both generation quality and multiview consistency (which also reflects camera controllability), highlighting the effectiveness of our designed model architecture. To demonstrate the advantages of our model in camera control performance more clearly, we evaluate our model on RealEstate10K, a dataset widely used in studies of single-view camera control generation. Our model shows better camera controllability by achieving a lower rotation and translation error, while maintaining comparable generation quality.

\subsubsection{Qualitative Results}
In Fig \ref{real10k}, we show the generation results on RealEstate10K dataset. Based on the given camera motion trajectory, our model can generate corresponding high-quality motion videos. Furthermore, comparisons with the baseline indicate that MyGo has advantages in both generation quality and camera controllability: as highlighted in the red boxes, the generated sofa in the video by CameraCtrl\cite{he2024cameractrlenablingcameracontrol} appears blurry, while the results of MotionCtrl\cite{motionctrl} can not follow input camera condition precisely. Fig \ref{main_result} shows the spatial and temporal consistency on nuScenes multi-view dataset. To more clearly demonstrate that our model retains camera control capabilities in multi-view scenarios, we present in Fig.\ref{newroute} an experiment where the vehicle changes lanes to the right. It can be observed that with the same road condition, our method allows for editing the ego car motion trajectory by editing the camera parameters.
\begin{table}[H]
\centering
\begin{tabular}{c|c|c|c}
\hline             
     \textbf{Model}   & FID↓ & FVD↓ & KPM Score↑  \\ \hline
     w/o camera injection& 12.88 & \textbf{117.50}&78.99\%\\ 
w/o epipolar mask&11.83  & 148.47&80.48\%  \\ 
                             \textbf{MyGo} (Ours) &  \textbf{11.60} &122.06   &\textbf{81.22\%}           \\  \hline
\end{tabular}
\caption{Ablation study of camera injection module and epipolar geometry constraints.}
\label{tab:ablation}
\end{table}

\subsection{Ablation Study}
Our method breaks down the task of generating multi-view driving videos with camera control into two parts: camera parameter controllability and multi-view spatial-temporal consistency. For the former, we inject plücker coordinates into the U-Net as described in Sec. \ref{subsec:3.3}. For the latter, MyGo applies camera pose-based epipolar constraints to the neighbor view cross-attention block as described in Sec. \ref{subsec:3.4}. Therefore, to verify the contribution of the two modules to achieving high-performance video generation, we conduct experiments in this session by removing each module individually. The quantitative results are shown in Tab. \ref{tab:ablation}.

To evaluate the contribution of camera control module purposed in Sec. \ref{subsec:3.3}, we ignore the camera plücker coordinates but preserve the part related with camera conditions in neighbour view cross-attention block. Without injecting camera conditions into U-Net, the model yield worse results on both FID and KPM Score. And model without plücker coordinates performs worse than camera control baseline methods on KPM Score, yet it is on par with the baseline in FID and FVD, representing generation quality, suggesting that the camera control module contributes more to motion control than to improving video quality.

In ablating the epipolar constraints, we simply remove the mask and allow complete cross-attention between latent feature and neighbour view conditions. All metrics show degradation, with the FVD score, which represents video quality, being the most adversely affected. The deterioration in metrics confirms that involving too many irrelevant regions in cross-attention process disrupts the generation and reduces consistency, and proves that our epipolar constraint is necessary to filter the overlapping regions in the neighbor views.

\section{Conclusion}
This paper introduces MyGo, the first method that integrating camera control into multi-view video generation. Injecting the plücker coordinates into the U-Net throug a ControlNet-like module preserves the generative capabilities of the pre-trained model while achieving precise camera control. We further use epipolar geometry based neighbour view cross-attention module to enhance multi-view consistency, which reinforces the content in the multi-view overlapping regions using prior information about the camera positions. Experiments on RealEstate10K and nuScenes shows that our method surpasses previous approaches in terms of camera controllability and multiview consistency.

\bibliography{aaai25}

\section*{Reproducibility Checklist}
This paper:
\begin{enumerate}
    \item Includes a conceptual outline and/or pseudocode description of AI methods introduced (yes)
    \item Clearly delineates statements that are opinions, hypotheses, and speculations from objective facts and results (yes)
    \item Provides well marked pedagogical references for less-familiar readers to gain background necessary to replicate the paper (yes)
\end{enumerate}
Does this paper make theoretical contributions? (yes)
\begin{enumerate}
    \item All assumptions and restrictions are stated clearly and formally. (yes)
    \item All novel claims are stated formally (e.g., in theorem statements). (yes)
    \item Proofs of all novel claims are included. (yes)
    \item Proof sketches or intuitions are given for complex and/or novel results. (yes)
    \item Appropriate citations to theoretical tools used are given. (yes)
    \item All theoretical claims are demonstrated empirically to hold. (yes)
    \item All experimental code used to eliminate or disprove claims is included. (yes)
\end{enumerate}
Does this paper rely on one or more datasets? (yes)
\begin{enumerate}
    \item A motivation is given for why the experiments are conducted on the selected datasets (yes)
    \item All novel datasets introduced in this paper are included in a data appendix. (yes)
    \item All novel datasets introduced in this paper will be made publicly available upon publication of the paper with a license that allows free usage for research purposes. (yes)
    \item All datasets drawn from the existing literature (potentially including authors’ own previously published work) are accompanied by appropriate citations. (yes)
    \item All datasets drawn from the existing literature (potentially including authors’ own previously published work) are publicly available. (yes)
    \item All datasets that are not publicly available are described in detail, with explanation why publicly available alternatives are not scientifically satisficing. (yes)
\end{enumerate}

\end{document}


\maketitle

%

\section{Implementation Details}
\subsection{Epipolar Constraints}
The epipolar geometry guided neighbour view cross-attention block(Sec. 3.5) takes latent feature of $\mathbb{R}^{frame,channel,h,w}$, concatenated neighbour view embedding of $\mathbb{R}^{frame,channel,h,2w}$ and camera poses $[\mathbf{R}, \mathbf{T}, \mathbf{K}] \in\mathbb{R}^{frame,7,3}$ as input. For local camera and neighbour camera poses, the fundamental matrix can be calculated as ($N$ denotes neighbour and $L$ denotes local view):
\begin{equation}
F_{N\to{L}} = K_N^{-1}R[t]_{\times}K_L^{-1}
\end{equation}
where $R$ and $t$ are the relalive rotation from neighbour view to local view:
\begin{equation}
    R = R_N^TR_L
\end{equation}
\begin{equation}
    t = R_N^T(t_L - t_N)
\end{equation}
When local camera id $L = i$ and neighbour camera id $N = i\pm1$, we can calculate the cross attention mask $Mask_{i\pm1\to{i}}\in\mathbb{R}^{h\times{w}, 2\times{h}\times{w}}$ as below:
\begin{equation}
Mask_{i\pm1\to{i}}(x_{i\pm1},x_i) = \mathbf{1}(x_iF_{i\pm1\to{i}}x_{i\pm1}^T < \tau)
\end{equation}
To ensure that in the neighbor view, only the most critical points are involved in cross-attention, we filter these points at a fixed ratio and determine the value of $\tau$, where we set the ratio as 0.25.
\subsection{Training Details}
To enhance the model's performance in autonomous driving scenarios, our implementation of integrating camera conditions, as well as other baseline methods on the nuScenes dataset, is based on multimodal inputs, including the first frame of each view, BEV map, and 3D bounding boxes. During the training process, we drop each condition with a 40\% probability, while specially we drop neighbour view condition with a 50\% probability considering during inerence there are only half of the views to which the neighbour view condition is available.

Our U-Net backbone consists of four downsample blocks, one mid-temporal transformer block, and four upsample blocks, with each block containing a single transformer layer. We set the video to  2 frames per second both in training and inference process.

\section{Limitation and future works.}
In the task of driving video generation, ego vehicle's motion is mostly in the forward and backward directions with significant displacement, which makes it challenging to apply  epipolar constraints or other rule-based methods for enhancing temporal-consistency. Using the first frame from each view as a condition provides some reference for the generation process, but it is still insufficient to handle scenarios with significant ego vehicle movement. Therefore, it is necessary to further investigate methods for enhancing temporal consistency.
\section{Supplementary Visualization Results}

\subsection{Results on RealEstate10K Dataset}
\begin{figure}[ht]
\centering
\includegraphics[width=\columnwidth]{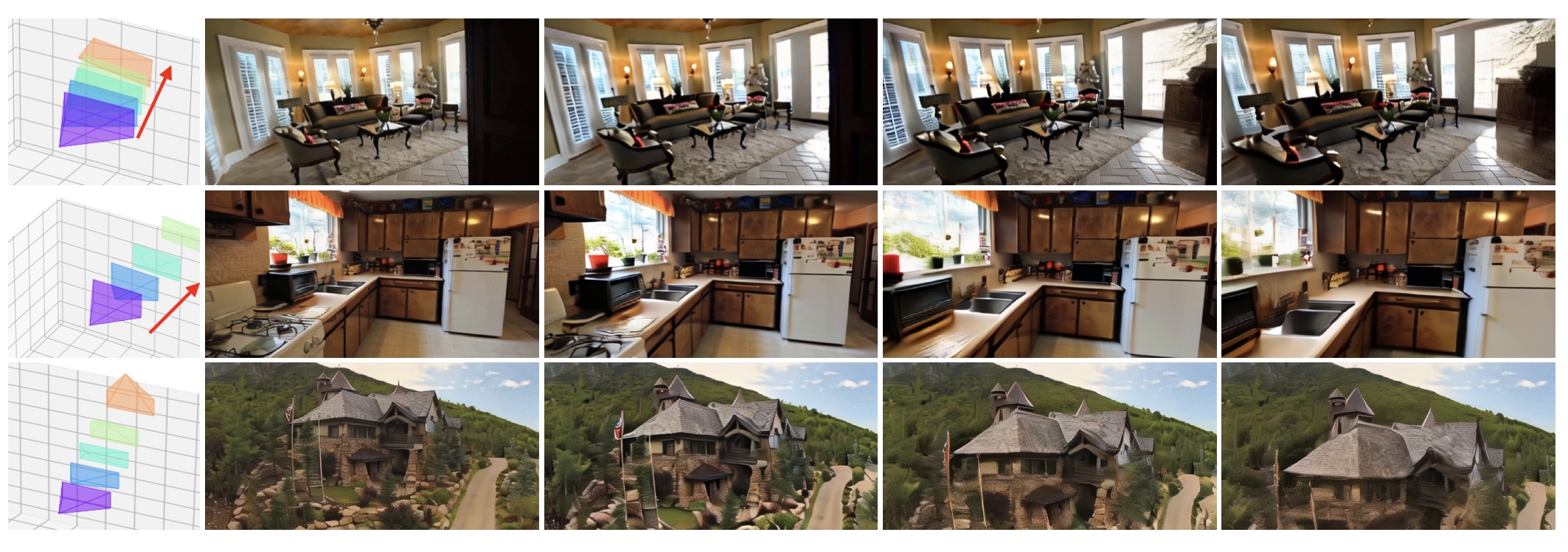} 
\caption{Examples of the generated camera control video frames on RealEstates10K. In the single-view camera control task, our approach also demonstrates excellent controllability and generation quality, which validates the effectiveness of our camera control module.}
\label{real10k_res}
\end{figure}

See in Fig. \ref{real10k_res}
\subsection{Results on Multi-View nuScenes Dataset}
\begin{figure*}[ht]
\centering
\includegraphics[width=1.8\columnwidth]{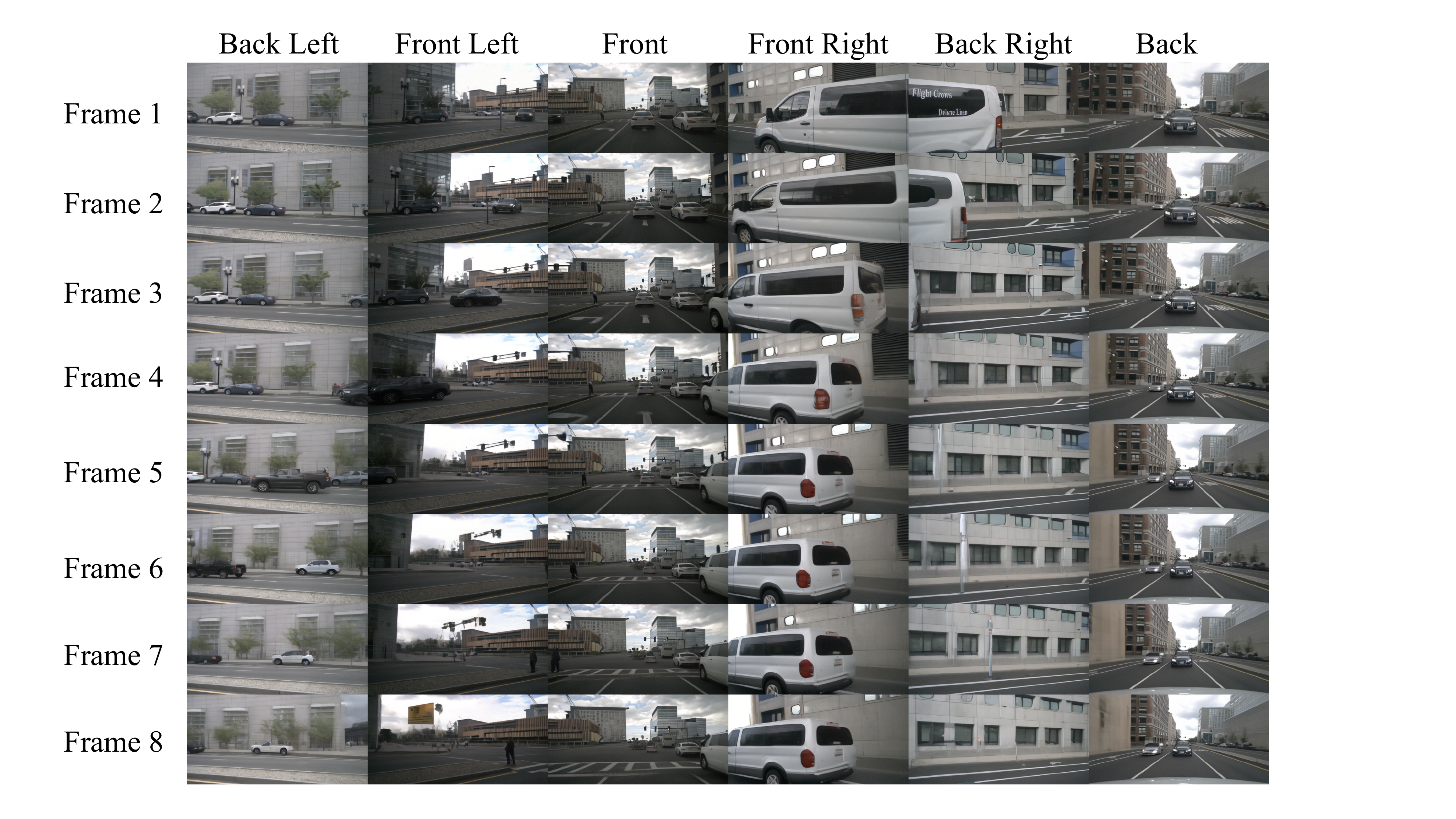} 
\includegraphics[width=1.8\columnwidth]{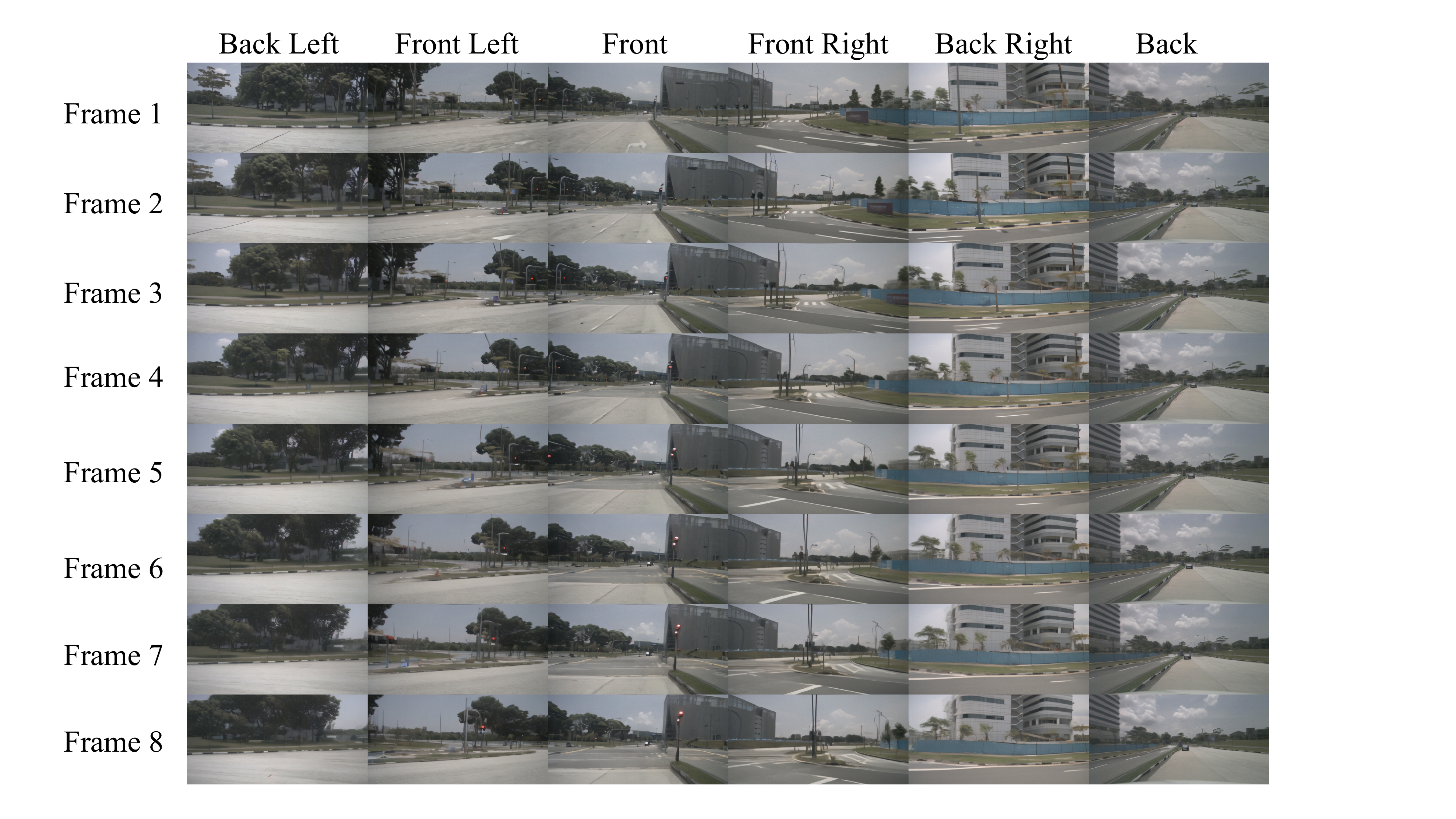} 
\caption{Examples of the generated multi-view video frames on nuScenes. Our method shows capability to generate videos with consistency and camera controllabilty}
\label{nuc_res}
\end{figure*}

See in Fig. \ref{nuc_res}

